\documentclass[sigconf,review,anonymous]{vldb}
\usepackage{graphicx}
\usepackage{balance}  
\usepackage{lipsum}             

\usepackage{multirow}
\usepackage{verbatim}
\usepackage{titlesec}
\usepackage{graphics}
\usepackage{adjustbox}

\usepackage{lipsum}

\titlespacing\section{0pt}{12pt plus 4pt minus 2pt}{0pt plus 2pt minus 2pt}
\titlespacing\subsection{0pt}{12pt plus 4pt minus 2pt}{0pt plus 2pt minus 2pt}
\titlespacing\subsubsection{0pt}{12pt plus 4pt minus 2pt}{0pt plus 2pt minus 2pt}

\begin{document}


\title{Membership Inference Attack for Beluga Whales Discrimination}

\numberofauthors{6} 

\author{
%
%
\alignauthor
Voncarlos M. Araújo
       \affaddr{Département d’Informatique, Université du Québec à Montréal  }       
       \email{}
\alignauthor
Sébastien Gambs
       \affaddr{Département d’Informatique, Université du Québec à Montréal  }
       \email{}
\alignauthor Clément  Chion
       \affaddr{Département des Sciences Naturelles, Université du Québec en Outaouais}
       \email{}
\and  
\alignauthor Robert Michaud
       \affaddr{Groupe de recherche et d’éducation sur les mammifères marins (GREMM)}
       \email{}
\alignauthor Léo Schneider 
       \affaddr{Département d’Informatique, Université du Québec à Montréal  }
       \email{}
\alignauthor   Hadrien Lautraite
       \affaddr{Département d’Informatique, Université du Québec à Montréal  }
       \email{}
}

\maketitle

\begin{abstract}
To efficiently monitor the growth and evolution of a particular wildlife population, one of the main fundamental challenges to address in animal ecology is the re-identification of individuals that have been previously encountered but also the discrimination between known and unknown individuals (the so-called ``open-set problem''), which is the first step to realize before re-identification.
In particular, in this work, we are interested in the discrimination within digital photos of beluga whales, which are known to be among the most challenging marine species to discriminate due to their lack of distinctive features. 
To tackle this problem, we propose a novel approach based on the use of Membership Inference Attacks (MIAs), which are normally used to assess the privacy risks associated with releasing a particular machine learning model.
More precisely, we demonstrate that the problem of discriminating between known and unknown individuals can be solved efficiently using state-of-the-art approaches for MIAs. 
Extensive experiments on three benchmark datasets related to whales, two different neural network architectures, and three MIA clearly demonstrate the performance of the approach.
In addition, we have also designed a novel MIA strategy that we coined as ensemble MIA, which combines the outputs of different MIAs to increase the attack accuracy while diminishing the false positive rate.
Overall, one of our main objectives is also to show that the research on privacy attacks can also be leveraged ``for good'' by helping to address practical challenges encountered in animal ecology.
\end{abstract}

\section{Introduction}

In animal ecology, the ability to re-identify (re-ID) an individual animal across multiple encounters allows for addressing a broad range of questions such as ecosystem function, community, and population dynamics as well as behavioral ecology~\cite{Arts2015, Krebs1999}.
In many cases, especially for aquatic species such as marine mammals, re-ID requires extensive training and practical experience for a human to acquire sufficient expertise to be able to accurately recognize a particular individual. 
To partially circumvent this issue, biologists usually rely on approaches such as tagging and photo-identification (photo-ID)~\cite{articlepratical, Krebs1999}. 
While accurate, the tagging approach is intrusive to animals and is often expensive and laborious. 
In contrast, the photo-ID approach uses visual identification from camera images (\emph{e.g.}, hand-held camera, camera trap, or drones), which is non-invasive for animals and has a lower cost.
Nonetheless, there are some practical and methodological challenges associated with its use. First, even among experienced researchers, there is a non-negligible chance of human error and bias when reviewing photos~\cite{foster2012}. 
Second, it is also time-consuming and expensive in terms of human involvement to manually filter through thousands of images.

To overcome these limitations, one possible strategy is to rely on computer vision techniques to standardize and automatize the animal re-ID process~\cite{Schneider2019}. 
To realize this, for decades, ``feature engineering'', which can be defined as the process of selecting or transforming raw data into informative features, has been the most commonly used technique.
Basically, it means that most of the algorithms for animal re-ID are designed and implemented to focus exclusively on predetermined traits, such as patterns of spots or stripes, to discriminate among individuals. 
However, feature engineering requires programming experience, sufficient familiarity with the species considered to identify relevant features.
In addition, this approach lacks in generality as once a feature detection algorithm has been designed for one species, it is unlikely to be useful for others~\cite{tiger}.

More recently, the last decade has witnessed the emergence of deep learning systems that make use of large data volumes to automatically learn discriminative features~\cite{7906512}.
In particular, Convolutional Neural Networks (CNNs) have achieved state-of-the-art results in a variety of uses cases based on the assumption of a closed world (\emph{i.e.}, a fixed number of classes/identities), 
However, CNNs are known to lack robustness when deployed in real-world classification/recognition applications, in which incomplete knowledge of the world during training result in unknown classes being submitted to the model during testing.
This corresponds for instance to the situation in which when used in the wild, the model will have to recognize individuals that it has not seen during training.


In marine ecology, one of the main challenges related to animal re-ID, such as wild whales, is the encounter of large populations in which there is frequently the addition of new individual appearing due to birth or migration, therefore creating an ``open-set'' setting~\cite{6365193} wherein the identity model must deal with ``classes'' (\emph{i.e.}, individuals) unseen during training.
Thus, a desirable feature for an animal re-ID approach is to have the ability to identify not only animals that belong to the catalog but also recognize new individuals (\emph{i.e.}, previously unknown animals). 
To address this issue, we investigate the use of Membership Inference Attacks (MIA), which is a form of privacy leakage in which the objective of the adversary is to decide whether a given data sample was in a machine learning model’s training dataset~\cite{DBLP:conf/sp/ShokriSSS17, yeom2018, salem2019, nasri, long, Chen_2021}. 
Knowing that a specific data sample was used to train a particular model may lead to potential privacy breaches if for instance this membership reveals a sensitive characteristic (\emph{e.g.}, being part of the cohort of patients having a particular disease or being a member of a vulnerable group). 
The gist of our approach is we could leverage on a MIA to discriminate whether a new beluga whale was present or not in the training set.
Then, this information can be used in the re-ID pipeline to take the decision to classify known individuals or to add a new entry in the catalog for an unknown individual.

To summarize, in this paper our main contribution is the proposition of a novel approach for whales discrimination through images (photo-ID), which relies on the use of MIAs.
In particular, one of our objective is to show that by drawing on the significant body of work on MIAs, it is possible to efficiently address the ``open-set'' vs ``closed-set'' problem.
To demonstrate this, extensive experiments have been conducted with three state-of-the-art MIAs that leverage different information produced by the model (\emph{i.e.}, prediction confidence, predicted and ground truth label, or both of them) as well as different attack strategies (neural network-based, metric-based and query-based attacks).
More precisely, we have performed a comprehensive measurement of the success of MIAs to address the open-set problem over two model architectures (ResNet50~\cite{resnet50} and DenseNet-121~\cite{densenet121}), three benchmark image datasets related to whales species (GREMM~\cite{Michaud2014}, Humpback~\cite{humpdata, Cheeseman2021} and NOAA~\cite{nooa}) along with three state-of-the-art MIAs, namely Yeom \emph{et al.}~\cite{yeom2018}, Salem \emph{et al.}~\cite{salem2019} and LabelOnly~\cite{Choo2020}, thus building a total of 36 attack scenarios.
In addition, previous works~\cite{DBLP:conf/sp/ShokriSSS17,salem2019,Choo2020} assume the leak information is more likely for machine learning models on the influence of overfitting, we ensure this assumption by evaluating overfitted and non-overfitted models while monitoring the false positive rate as recommended in~\cite{carlini,onThediff} for the reliability of the results.
Finally, we introduced a novel attack design for whale discrimination, which we coined as ensemble MIAs, which combines the outputs of different MIAs to increase the attack accuracy while decreasing the false positive rate.

The outline of the paper is as follows.
First in Section~\ref{sect_related}, we review the relevant background on automated photo identification systems as well as on membership inference attack.
Then in Section~\ref{sect_approach}, we describe the St. Lawrence beluga whale re-id pipeline from side pictures, the training of the attack model as well as the different MIA strategies that we propose to implement the discrimination between known and unknown belugas.
Afterwards in Section~\ref{sect_experiments}, we present the experimental setting used to evaluate our approach, which includes the datasets, the experimental configuration as well as the target and attack models.
Finally in Section~\ref{sect_result}, we report on the performance of the approach under different scenarios, before discussing how the attack can generalize to different settings as well as the factors influencing its success and its robustness before concluding in Section~\ref{sect_conc}.

\section{Related Work}
\label{sect_related}

In this section, we first review the related work on re-identification and discrimination of marine mammals as well as the background on MIAs.

\subsection{Automated Photo Identification of Marine Mammals}

The research on the individual identification of cetaceans using natural markings began in the early 1970s \cite{review1,review2}, including the use of unique markings and coloration, or notches in the dorsal fin or fluke \cite{finreview,review3}. 
For instance, Pollicelli, Coscarella and Delieux~\cite{Pollicellire} have evaluated the opportunity to use image metadata (\emph{i.e.}, annotations describing the animal characteristics as well as the time and place at which the picture was taken) as an attribute for photo-ID to reduce the number of possible matches in the identification step. 
In this work, classical machine learning techniques, such as neural networks, Bayesian classifiers, decision trees and $k$-nearest neighbors, were applied on the metadata of 869 pictures taken of 223 Commerson’s dolphin
individuals taken over seven years. 
Overall, the decision tree classifier was able to correctly identify 90\% of the individuals on the validation set based only on the metadata of their pictures. 
One clear limitation of this work is the reliance on metadata rather than on intrinsic visual characteristics of the animals.
In addition, manual work is also required and the system has to be retrained to include new individuals. 

In~\cite{RENO201995}, a fully automated system called Smart Photo-ID of Risso’s dolphin (SPIR) was developed to study the presence of Risso’s dolphin in the Gulf of Taranto. 
This species is characterized by several distinctive scars over the dorsal fin present in the animal, a useful pattern for automated recognition. 
The dataset necessary for training this system was created with the general public involvement in research activities, side by side with experts. 
The first step of the system consists in preprocessing the input image to extract the dorsal fin segmentation employing Otsu’s  threshold technique \cite{otsu} and morphological operators. Followed by detection, feature extraction is performed using Speeded Up Robust Feature (SURF)~\cite{surf} and Scale-Invariant Feature Transform (SIFT)~\cite{sift}, which are methods for extracting local characteristics in images.
To predict the identity of an unknown dolphin, the input image is compared with all of the images available in the database.
Then, the picture with the highest number of matching features with the query image is selected as the best-matching dolphin. 
The results obtained demonstrate that SIFT outperforms the SURF feature detector, showing better performances and achieving a 90\% accuracy in the validation experiment. 
Unfortunately, the application of SPIR cannot be extended easily to other species, especially if these are not characterized by scars over the dorsal fin. 

Recently, Maglietta and collaborators~\cite{Maglietta2020} have proposed a novel methodology called NNPool, dedicated to the automatic discrimination of unknown vs. known Risso’s dolphins. 
More precisely, NNPool consists of a pool of $n$ CNNs, each one being trained to recognize a particular known individual versus the rest of the dolphin (\emph{i.e.}, a form of one-versus-all classification).
The models were trained on Risso’s dolphins data and photos acquired between 2013-2018 in the Northern Ionian Sea (Central-Eastern Mediterranean Sea). 
The results obtained have also been validated using another dataset composed of unknown images of Risso’s dolphins from the Northern Ionian Sea and the Azores, acquired in 2019. 
More precisely, their experiments considered 28 individuals to validate experimental results containing 300 images of Risso’s dolphin fins detailed as 40 images belonging to some of the 23 known dolphins with the the remaining 260 belonging to the unknown dolphins. 
The discrimination accuracy of 87\% was measured on a validation dataset, which can be used as preprocessing of SPIR~\cite{RENO201995} to detect an unknown dolphin before performing the photo-ID of known individuals. 
This work is the closest to our work, in the sense that it considers the discrimination task of distinguishing known vs unknown individuals, rather than only photo re-id.
Nonetheless, it is applied on a species that is much more easier to discriminate because of its distinctive marks. 
In addition, the dataset used to conduct the experiments is not publicly available, which makes it impossible to compare our approach to theirs.

Deep learning approaches are relatively novel in the field of animal photo-ID~\cite{chim, Miele2021, Korschens2019, Nepovinnykh2020, Maglietta2020}. 
For example, Bogucki and co-authors introduced a fully automated system based on three CNNS for photo-ID of North Atlantic right whales~\cite{Bogucki2019}. 
This system participated on the Kaggle platform in 2015, on the automation of the right whale recognition process using a dataset of aerial photographs of animals~\cite{humpdata}.
The training dataset provided for the competition consisted of 4544 images containing only one single right whale and were labeled with a correct whale. 
Submissions were evaluated on a test set of 2493 images, used to determine the rankings of the competitors. 
The numbers of pictures per whale varied considerably in this dataset (\emph{e.g.}, six individuals had only one photograph whereas there were two whales with eighty-two images each). 
This is a challenging setting for classification, whose performance depends on the number of images available for each individual. 
The proposed method uses a CNN that selects the region of interest 
and outputs a bounding box around the head of the whale, which is then used to crop the high-resolution image. 
The authors developed a network that automatically scales, rotates and crops the input image. 
This is achieved by training a CNN to locate two key points on the top of the whale’s head from already labeled data.
Data augmentation was applied, adding rotated and re-scaled versions of the images in the original dataset. 
Finally, another CNN was used to perform actual whale identification, obtaining an accuracy of individual right whale recognition of 87.44\%. 
The authors explained that the wide variability in the number of images per individual whale impacted the performance of the last CNN devoted to individual recognition. 
More precisely, having more images per individual improves the recognition accuracy.

More recently, Bergler and co-authors~\cite{Bergler2021} have developed a deep-learning-based framework for identifying killer whales. 
The approach, called FIN‑PRINT, was trained and evaluated on a dataset collected over an 8‑year period (2011–2018) in the coastal waters of western North America that consists of 367 individuals.
First, object detection is performed to identify unique killer whale markings, which results in 94.1\% of precision using the recent version of YOLO (YOLOv5) \cite{glenn_jocher_2022_7347926}. 
Second, all previously detected killer whale markings are extracted.
The third step introduces a data enhancement mechanism by filtering between valid versus invalid (VVI) markings from previous processing levels, in which ResNet34 is used for binary classification between VVI identification images achieving 97.5\% of precision. 
The fourth and final step involves multi‑class identification, which assigns a label to the top 100 killer whales for a test sample.
FIN‑PRINT achieves an accuracy of 92.5\% and 97.2\% using respectively top-1 and top‑3 for photo‑identified killer whales. 
Note that the top-100 killer whales each have more than 325 images per individual while the remaining individuals have fewer images per class, which leads to an unbalanced dataset challenge.

In Cheeseman \emph{et al.}~\cite{Cheeseman2021}, the authors have developed a new CNN-based similarity algorithm for humpback whales 
individuals. 
The method relies on a Densely Connected Convolutional Network (DenseNet) to extract key-points of an image of the ventral surface of the fluke and then train the CNN model. 
The extracted features are then compared against those of the reference set of previously known humpback whales for similarity. 
The Arc-Face algorithm~\cite{Deng_2021} uses fluke shape, edge pattern and surface markings to locate images in a hyper-sphere space in which proximity becomes the similarity measure. 
For testing, they evaluated the complete dataset of 15494 humpback whale individuals considering 33321 whale fluke images used in Kaggle competition.
The authors argues that CNN-based image recognition is much faster and more accurate than traditional manual matching, reducing the time for identifying a picture by over 98\% and decreasing the error rate from approximately 6–9\% to 1–3\%. 




 To the best of our knowledge, there is no automatic photo-ID systems for beluga whales re-identification available in the literature, due to individual challenges as they often lack unique or permanent pigmentation.
 In addition, they also do not have a dorsal fin, a feature common to other ice-inhabiting whales (\emph{e.g}, humpback whales). 
 Although photo-ID studies of beluga whales are being conducted in Cook Inlet, Alaska~\cite{Mcguire}, the White Sea, Russia~\cite{popu}, and the St. Lawrence Estuary, Canada~\cite{Michaud2014}, a standardized and public database is not yet available for this task to be investigated by the computer vision scientific community. 

\subsection{Membership Inference Attack}

With respect to privacy, in addition to the sensitive inferences that can be drawn from the data itself, it is also important to understand how much the output of the learning algorithm itself (\emph{e.g.}, the model) leaks information about the input data it was trained on. 
For instance, privacy attacks (also called inference attacks) have been developed against machine learning models to reconstruct the training data from the model or to predict whether the profile of a particular individual known to the adversary was in the training dataset~\cite{DBLP:conf/sp/ShokriSSS17}. 
Generally, this membership inference is deemed problematic if revealing that a profile belongs to this database enables you to learn a sensitive information about this individual (\emph{e.g.}, the training set is composed of individuals suffering from a particular disease or from particularly vulnerable subgroups).

More precisely, in a MIA an adversary that knows the particular profile of an individual tries to infer whether this profile was in the training dataset used to learn a particular model~\cite{Hu2022}.
Generally, the adversary models considered in the MIA literature assume either a black-box or white-box access to the model being attacked.
In a black-box setting, the term oracle is sometimes used to refer to the access of the adversary, since he can only submit requests to the model and observe the model outputs (\emph{i.e.}, he does not have access to the structure of the model).  
Such attacks need little information and as such are quite general and versatile but at the same time offer usually lower performances than attacks conducted in the white-box setting. 
In contrast, a white-box adversary is assumed to have a (partial or full) knowledge of the model such as its architecture, its parameters as well as its weights. 
The attacks conducted in this setting usually achieve better performances since they can be adapted to specific models and also have access to more information at inference time. 

Usually, the success of MIA attacks is impacted by model overfitting. 
Indeed, if the model attacked has overfitted the training data, it will behave quite differently when an example contained in the training set is submitted (\emph{e.g.}, the confidence on its prediction will be higher). 
This means that the success of MIAs can be decreased by employing mechanisms classically used in machine learning to reduce overfitting as well as by using more training samples to avoid too precise memorization.
In contrast in our case, we will exploit overfitting in a positive way as a manner to increase the success of the MIAs and thus the discrimination of known vs unknown belugas.

Standard machine learning metrics such as precision, recall and F1-measure can be used to quantify the success of MIAs. 
However, they might be interpreted differently. 
For instance, in the attack context, one might want to have a high precision even if it means to reduce recall (\emph{e.g.}, by tuning a confidence threshold). 
Indeed, realizing a MIA on few individuals with a high confidence can be considered a higher privacy breach than performing a MIA on a large number of individuals but with a low confidence~\cite{carlini}.
More precisely, the false positive rate (also called sometimes the false alarm rate) should be reduced as much as possible to be a good indicator of an attack performance.

\section{Proposed Approach}
\label{sect_approach}

In this section, we first describe the generic beluga whale re-id pipeline before detailing the training process for the attack model as well as the different MIAs from the state-of-the-art that we have used to implement it.
Finally, we also describe our novel approach to perform MIA based on an ensemble strategy.

\subsection{Beluga Whale Identification Pipeline}

\begin{figure}[h!]
  \centering
  \includegraphics[width=\linewidth]{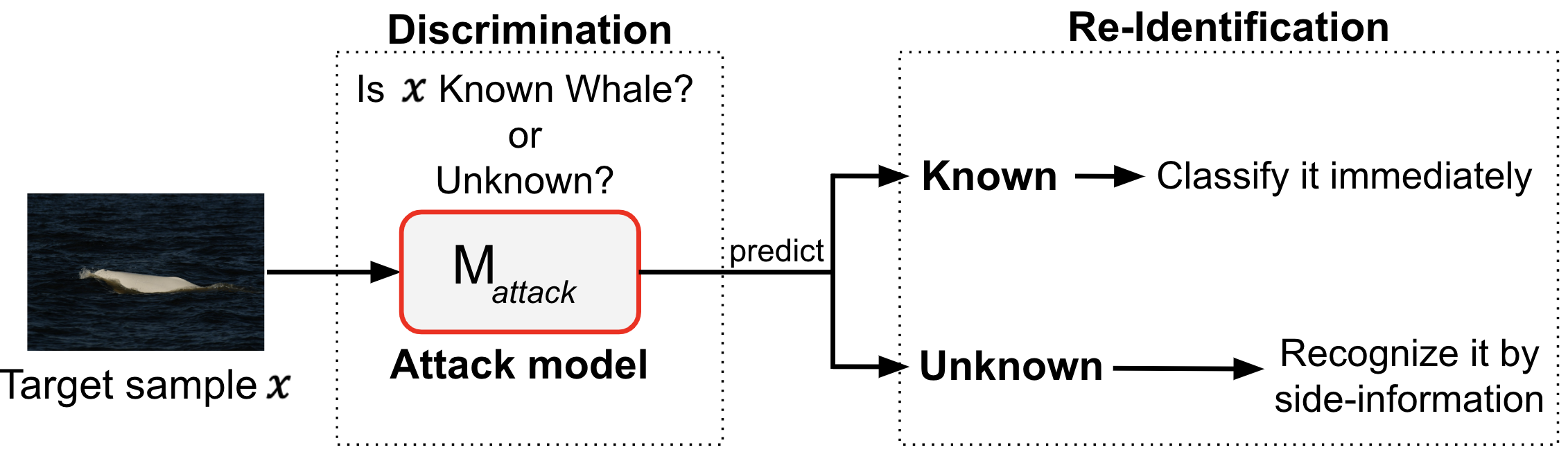}
  \caption{Overview of the whale identification pipeline.}
  \label{inferenceAttack}
\end{figure}

The general pipeline for beluga whale identification pipeline is illustrated in Figure~\ref{inferenceAttack}. 
It consists of two phases: (1) discrimination for distinguishing between known and unknown whale individuals through a MIA and (2) re-identification (re-ID).

\textbf{Discrimination.} The attack model trained to conduct the MIA is used to determine whether the target sample $x$ of a beluga is part of the training set of the target model. 
We describe how to build the attack model in Section~\ref{attackModeltrainSection}.

\textbf{Re-ID.} Once it has been determined whether the target sample $x$ corresponds to a known (\emph{i.e.}, within training set) or unknown beluga (\emph{i.e.}, out of training set) by the attack model, known examples can be immediately classified through a standard classifier. 
Otherwise for unknown belugas, the recognition has to be done manually through side-information. 
For instance, this side-information could be acquired through experts to confirm whether that individual is indeed new and otherwise to decide to which class the unknown individual will be assigned.

\subsection{Training of the Attack Model}
\label{attackModeltrainSection}

\begin{figure*}[h!]
    \centering
    \includegraphics[width=0.85\textwidth]{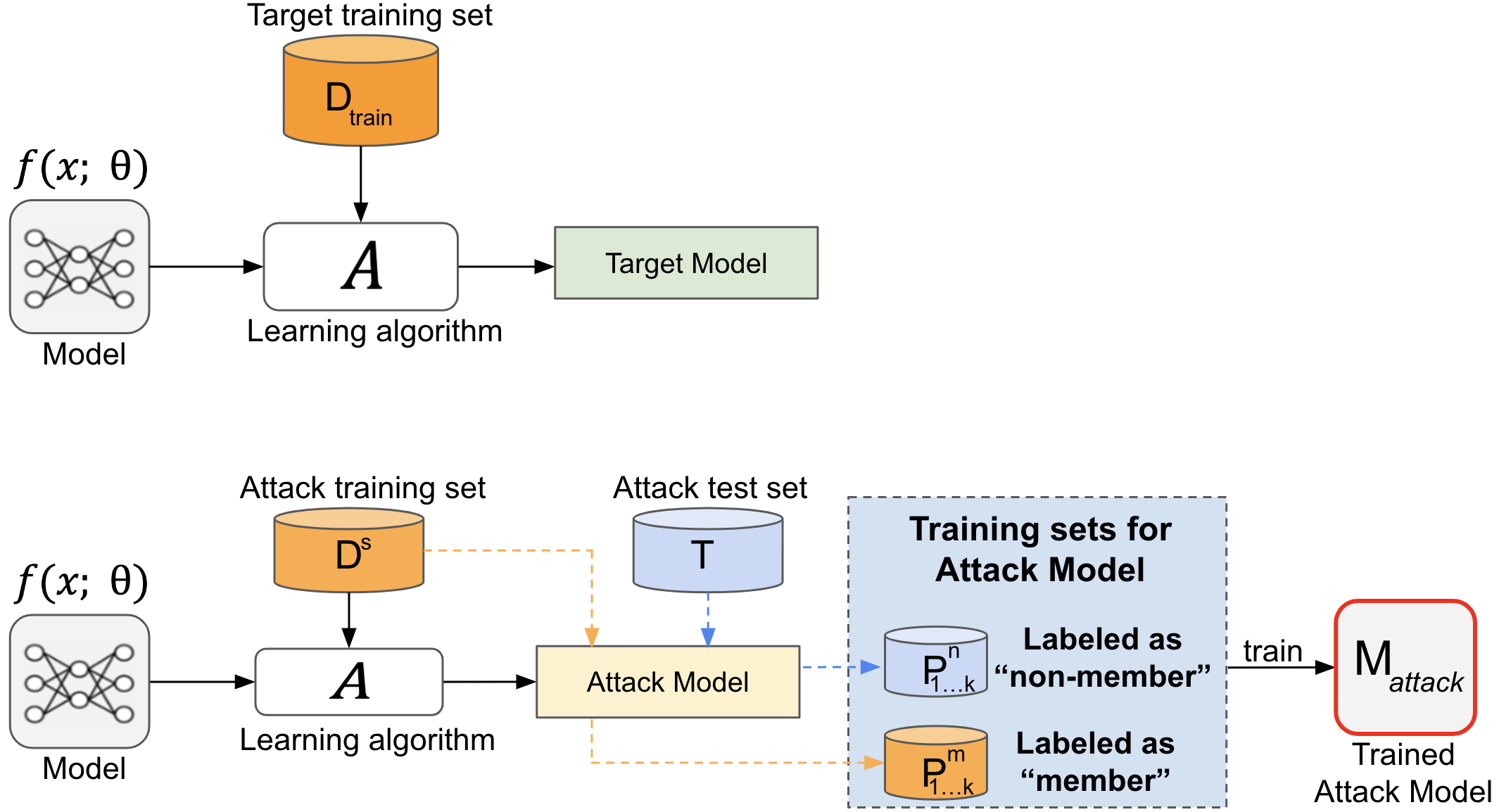}
    \caption{Attack model training procedure.}
    \label{trainAttack}
\end{figure*}

We assume that the adversary has access to a local dataset, which we call the attack dataset $D^s$. 
The attack dataset comes from a same distribution (\emph{i.e.}, the same population or the same individuals) than the one used to train the target model. 
To infer whether the sample $x$ is in the training set of the target model, our core idea is to train an attack model $M_{attack}$ that can detect whether a particular sample corresponds to the picture of a beluga that was part or not of the training set. 
Figure~\ref{trainAttack} provides a high-level overview of the training process of the attack model, which we describe in details hereafter.

\textbf{Training process.} 
$D_{train}$ is the training dataset, which is used for training the target model using the learning algorithm $A$. 
$D^{s}$ is the attack dataset that is disjoint from the training dataset $D_{train}$, which contains data points coming from the same data distribution as the training members in $D_{train}$. 
The adversary first trains the attack model using the attack training dataset $D_{s}$ and the learning algorithm $A$, in such as way that the attack model mimics the behavior of the target model. 
$T$ is the attack test dataset that is assumed to be both disjoint from $D^{s}$ and $D_{train}$, in the sense that it is composed by non-member individuals never seen before by $D^{s}$ and $D_{train}$. 
When the training of the attack model is completed, the adversary queries the attack model using the attack training and test datasets to obtain the outputted prediction vectors for each data point.
More formally, we denote a prediction vector as $\hat{p}(y\, |\, x)$, in which ``member'' are labelled as 1 and ``non-member'' as 0.
Then, each ``member'' dataset and ``non-member'' dataset are represented as follows:
\begin{equation}
\label{membAndnonMemb1}
P_{i}^{m} = \left\{\hat{p}(y \, | \, x),1\right\}
\end{equation}
\begin{equation}
\label{membAndnonMemb2}
P_{i}^{n} = \left\{\hat{p}(y \, | \, x),0\right\}
\end{equation}
More precisely, the prediction vector of each point $i$ in the attack training dataset is labeled ``member'' $P^{m}_{1}, \ldots, P^{m}_{k}$ and the prediction vector of each point $i$ in the attack test dataset is labeled ``non-member'' $P^{n}_{1}, \ldots, P^{n}_{k}$. 
Thus, the adversary can build $k$ ``member'' data points and $k$ ``non-member'' datapoints, which jointly form the training dataset of the attack model.

Finally, the problem of recognizing the complex relationship between members and non-members is converted into a binary classification problem. 
Once trained, the adversary can use the attack model $M_{attack}$ to implement MIAs on arbitrary data points. 
The attack model takes the prediction vector $\hat{p}(y\, |\, x)$ of the target model of a data point $x$ as input and outputs whether this point is in $D_{train}$ of the target model or not. 


\subsection{Attack Model Design}

To instantiate the MIA attack, we have applied three state-of-the-art MIAs from the literature that leverage different types of information outputted by the target model (namely prediction confidence, ground truth label or both of them) and different attack strategies (\emph{i.e.}, neural network-based, metric-based and query-based) as shown in Table~\ref{miafeature}.
These MIAs all consider an adversary with a black-box access to the model and thus are quite generic.
Note that while we did not consider MIAs with a white-box access to the model~\cite{white}, we leave as a future work their investigation to increase the attack success of beluga discrimination.
In security, white-box access is usually considered less realistic than black-box access in many real-life situations (\emph{e.g.}, the use of a machine learning as service). 
However, this is not the case in our setting as we can fully control the implementation pipeline of the MIA.

\begin{table}[h!]
\centering

\caption{Summary of MIAs investigated. 
In the features column, C denotes the use of confidence while L corresponds to the use of label.}
\label{miafeature}
\begin{tabular}{|c|c|c|} \hline

MIA   &         Features     &  Attack strategy \\ \hline
Yeom \emph{et al.}~\cite{yeom2018}   &     C, L            & Metric-based           \\ \hline
Salem \emph{et al.}~\cite{salem2019} &             C             & Neural network-based             \\   \hline

Label-only~\cite{Choo2020}   &  L              & Query-based                        \\ \hline

\end{tabular}
\end{table}


\textbf{Yeom \emph{et al.}~\cite{yeom2018}} In contrast to neural network-based attacks like Salem \emph{et al.}~\cite{salem2019} explained in the next paragraph, metric-based attacks leverage a certain metric and a predefined threshold over the metric (computed over the attack dataset by querying the attack model) to differentiate members and non-members. 
More precisely, Yeom \emph{et al.}~\cite{yeom2018} uses the prediction confidence of the correct class under the assumption that the confidence should be high for the member samples as the target model is optimized with this objective.
The $Metric_{conf}$ attack is defined as follows:

\begin{equation}
\label{yeomloss}
Metric_{conf} (\hat{p}(y \, | \, x), \, y) = -(\mathcal{L}(\hat{p}(y \, | \, x); \, y) \leq \tau),
\end{equation}
in which $\mathcal{L}$ is the cross-entropy loss function and $\tau$ is a preset threshold. 
An adversary infers an input record as a member if its prediction loss is smaller than the average loss of all training members while otherwise, it is inferred as a non-member. 
The intuition for this attack is that the target model has been learnt on its training members by minimizing their prediction loss. 
Thus, the prediction loss of a training record should be smaller than the prediction loss of a test record. 
The threshold is an input hyperparameter to the attack and as such he could be learned by using an evaluation set for instance. 
To identify the threshold for optimal accuracy, we use the evaluation set from the target set and treat one-half as members, with the rest as non-members. 
We compute the AUC (Area Under the Curve) and precision/recall metrics, sweep over a range of values for the threshold $\tau$ and measure the resulting attack’s FPR/TPR and precision/recall trade-offs. 
We can then choose the best threshold $\tau$ based on membership inference accuracy for this simulated setup.


\textbf{Salem \emph{et al.}~\cite{salem2019}} This attack takes the prediction vector confidences as the input to the attack model. 
The adversary derives the training dataset of the attack model by querying the attack model with the attack training dataset (labeled as members) and attack testing dataset (labeled as non-members). 
With the attack training dataset, the adversary can learn the attack model, which is a multi-layer perceptron (MLP). 
A traditional 3-layer MLP with 64, 32 and 2 hidden neurons for each layer is used for neural network-based attacks. 
We use the same hyperparameters of overfitting setting as seen in Section~\ref{sec2}.
Once the attack model is learnt, the adversary can perform the attack over the target model to differentiate members and non-members with respect to $D_{train}$.

\textbf{Label-only.~\cite{Choo2020}} 
Rather than using confidence predictions, query-based attacks restrict the attack to using only the predicted labels from the target model. 
Label-only attacks determine membership status by sending multiple queries to the target model, which concretely are generated by adding adversarial perturbations to the input sample until the predicted label has been changed. 
The attack measures the magnitude of the perturbation and considers the data sample as a member if its magnitude is larger than a predefined threshold.
More formally, given some estimate $dist(x, y)$ of a point’s $l2$-$distance$ to the model’s boundary, the attack predict $x$ a member if $dist(x, y) \geq \tau$ for some threshold $\tau$.

To estimate the distance, the attack starts from a random point $x$, which  is misclassified. 
Then, a ``walk'' along the boundary is performed while minimizing the distance to $x$ using HopSkipJump~\cite{chennmia}, which closely approximates stronger white-box attacks. 
HopSkipJump is initialized with a sample blended with uniform noise that is misclassified over iterations by moving it along the decision boundary to get closer to the attacked image.
For a given target model, the attack assumes that the robustness to adversarial perturbations is higher for a member sample compared to a non-member as the former was involved in the training of the model.



\subsection{Ensemble Membership Inference Attack}
\label{sec_ensemble_mia}

\begin{figure}[h!]
  \centering
  \includegraphics[width=\linewidth]{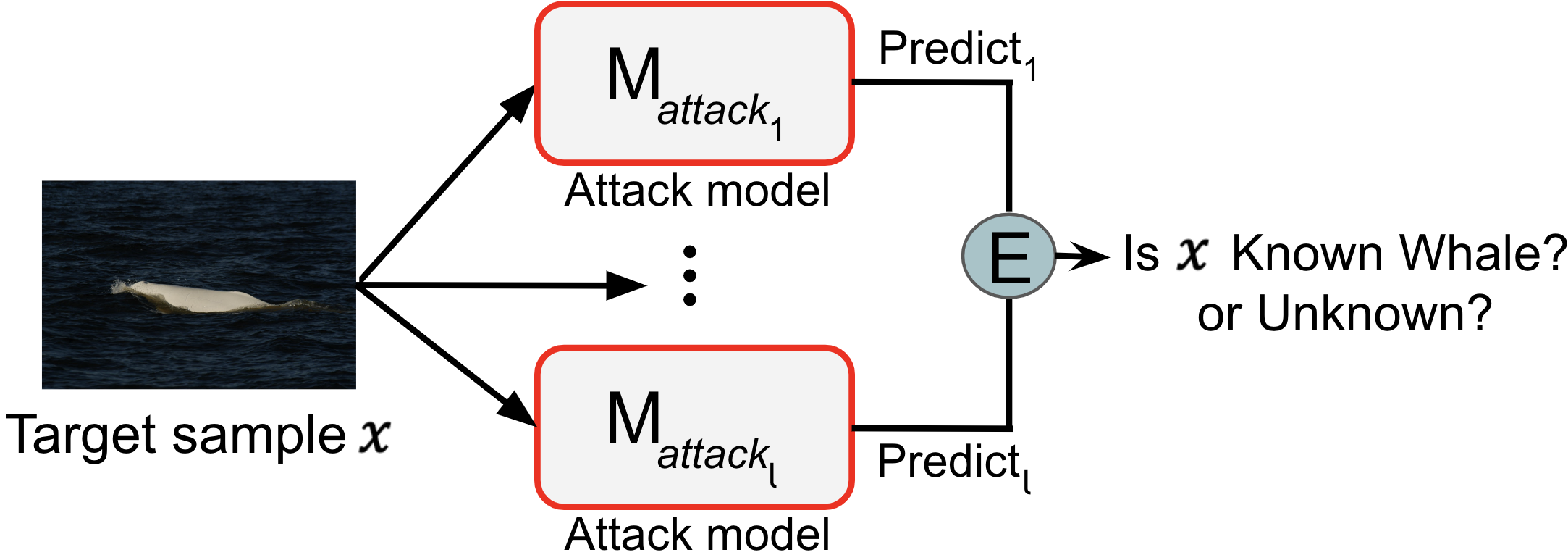}
  \caption{Ensemble MIA.}
  \label{EnsembleAttack}
  
\end{figure}

In this section, we propose a novel way to perform a MIA illustrated in Figure~\ref{EnsembleAttack}, which we coin as a ensemble membership inference attack. 
In the ensemble MIA, instead of a single one, $n$ attack models are built using different subsets of the data. 
More precisely, the attack model $M_{attack}$ is not trained directly using the whole dataset, but rather this dataset is split in disjoint subsets to create several attack models ($M_{attack_{1}}$ to $M_{attack_{l}}$).
For instance, when the dataset contains 60 individuals, it is split in 6 subsets with 10 individuals to train the $M_{attack_{1}}$ to $M_{attack_{6}}$.

At discrimination time, the MIA ensemble generates $l$ predicted outputs for each new sample $x$ that are combined using a combination rule $E$. 
The combination rule generates the final output ``member'' or ``non-member''. 
In this paper, we have used the simple combination rule $E$ that an input $x$ is labelled as ``member'' if at least one $M_{attack}$ prediction output assigned it as a ``member'', while otherwise it is considered to be a ``non-member''. 
Our rationale behind the design of the ensemble MIA is that training an attack model with fewer individuals makes the classifier more powerful to discriminate between similar classes. 
Indeed, smaller subsets decrease the complexity of discrimination as usually they give a higher prediction score for individuals seen during training that models built on bigger dataset. 
Moreover, it was confirmed in the experiments that the attack performance may vary across different individuals due to the different overfitting levels for each set of classes (see Table~\ref{Overffitingclasses} in Section~\ref{reliabilityM}). 




\section{Experimental Setting}
\label{sect_experiments}

In this section, we present the experimental setting used to validate our approach of using MIA for beluga whale discrimination. 
More precisely, we first describe the datasets used in the experiments (Section~\ref{datasets1}), followed by the experimental configuration (Section~\ref{sec_exp_configuration}) and finally by the target and attack models’ architectures and training settings (Section~\ref{sec2}).

\subsection{Datasets}
\label{datasets1}

The experiments were conducted on three distinct datasets in terms of visual characteristics: GREMM, Humpback and NOAA.
\begin{itemize}
\item The GREMM dataset is made of photos from hand-held cameras taken during photo-identification surveys conducted from June to October between 1989 and 2007 as part of an ongoing long-term study of the social organization of the St. Lawrence Estuary beluga population in Québec (Canada). 
This dataset is composed of 983 beluga individuals and thousands of side-view beluga images.
However, the number of pictures per individual varies significantly with a lot of belugas having only a small number of pictures.
Thus, we selected a part of this dataset that contains 3402 images distributed across 180 individuals. 
In addition, as a pre-processing step, we use the method previously proposed in~\cite{Araujo2022} to detect and crop the images. 
\item The Humpback dataset was derived from the Happywhale - Whale and Dolphin Identification competition dataset~\cite{humpdata}, which originally contains images of over 15000 unique individual marine mammals from 30 different species collected from 28 different research organizations. 
We selected just the Humpback species to evaluate our approach because they are known to be among one of the easiest species to recognize due to their very distinctive patterns on the flukes.
For example, the first ranked solution~\cite{1placeKagl} in Kaggle competition achieves 0.973 on the private leader-board in a competition that only identifies humpback whales~\cite{Simoes2020}.
The Humpback dataset contains 270 individuals with a total of 4814 images.
\item Finally, the last dataset has been collected by the US federal agency National Oceanic and Atmospheric Administration (NOAA), which monitors five different populations of belugas across Alaskan waters, with a focus on the Cook Inlet belugas. 
More precisely, the NOAA dataset is composed by 380 individuals with 5158 images in total corresponding to top-view from belugas whales.
Note that the top-view pictures that compose the NOAA dataset are considered more informative than pictures of beluga flanks taken from the side of animals that compose the GREMM dataset.
\end{itemize}

\begin{figure}[h!]
  \centering
  \includegraphics[width=\linewidth]{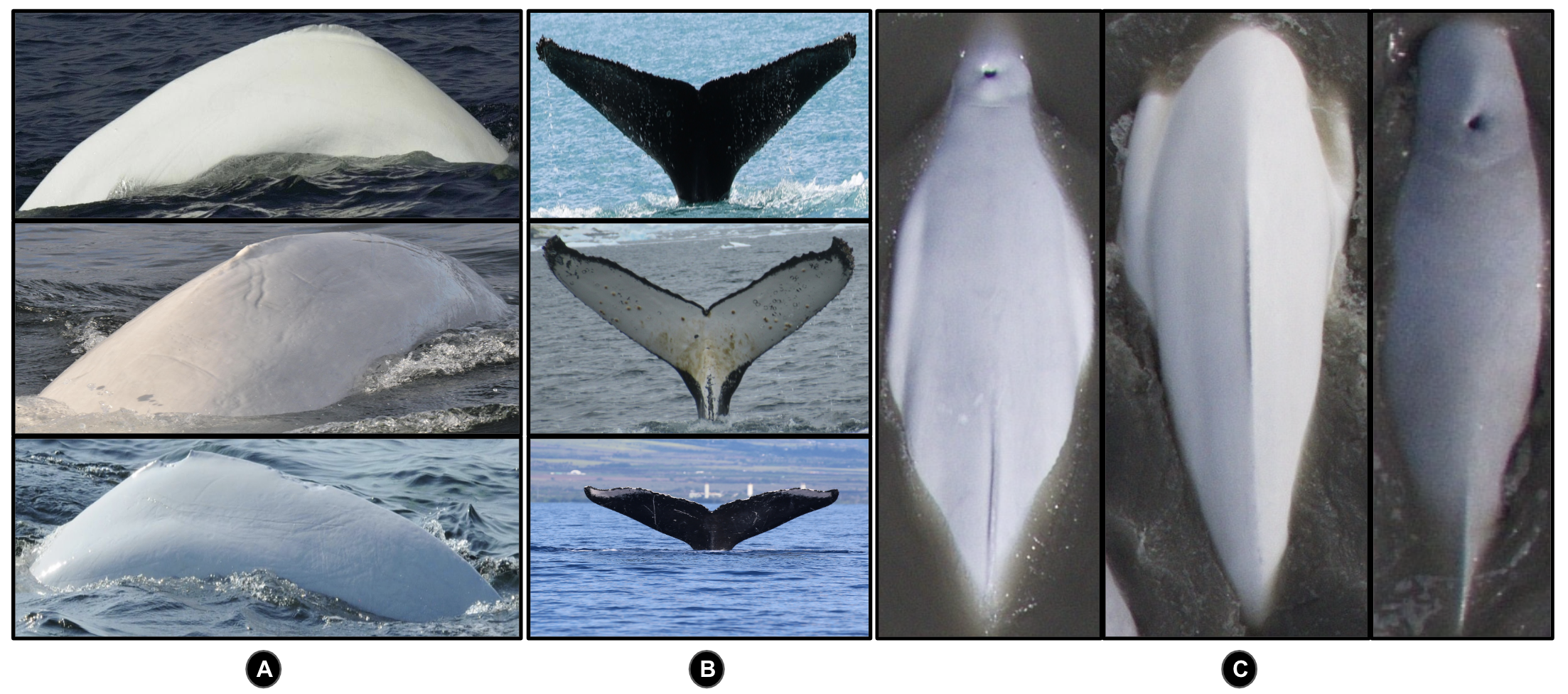}
  \caption{Visual characteristics of datasets. (A) GREMM, (B) Humpback and (C) NOAA.}
  \label{datasetImage}
\end{figure}

To summarize, Table~\ref{tableDataset} describes the numbers of individuals and sample pictures in each of these three datasets while their visual characteristics are presented in Figure~\ref{datasetImage}. 

\begin{table}[h!]
\centering
\caption{Statistical information of datasets.}
\label{tableDataset}
\begin{tabular}{|c|c|c|} \hline
Dataset   &         Nb of individuals     &  Nb of samples \\ \hline
GREMM  &             180             & 3402             \\   \hline

Humpback   &     270            & 4814            \\ \hline
NOAA &  380             & 5158                        \\ 
\hline
\end{tabular}
\end{table}

\subsection{Experimental Configuration}
\label{sec_exp_configuration}

Each dataset used to train MIAs is composed of individual whale, each one of them having many pictures associated to it. 
To assess the MIA for beluga discrimination, we have sampled three disjoint sub-datasets with equal or approximately equal number of identities. 
However, the number of images per individual (\emph{i.e.}, an individual beluga is an identity) is very diverse from one beluga to another. 
Therefore, the sampling cannot guarantee that each class has an equal number of data points in each sub-dataset unless increasing the number of data points in each sub-dataset. 
The augmentation process included operations such as random horizontal flip and brightness variation. 
More precisely, augmented images are obtained by rotating the original image 90, 180, 270 and 330 degrees clockwise and random brightness between 0.0 and 1.0~\cite{augbloice}.
Based on this observation, we construct three disjoint augmented subsets to which we assign randomly the identities of belugas: the target set, the attack set and the evaluation set (see Table~\ref{tableXXX}). 
After augmenting for those individuals that contains fewer samples, this leads to 75 images per ID. 
Thus, each of this dataset will contain approximately $\frac{1}{3}$ of the identities ($ID^1$, $ID^2$ and $ID^3$). 
More precisely, the GREMM, Humpback and NOAA datasets are composed respectively of 60, 90 and 127 individuals with a number of images per subset of respectively 1500, 2250 and 3175 images for GREMM, Humpback and NOAA datasets respectively. 
In total, the evaluation set ($ID^1$ and $ID^3$) is composed of 3000, 4500, and 6350 samples for respectively GREMM, Humpback and NOAA datasets.

\begin{table}[h!]
\caption{Summary of experimental configuration datasets.
}\label{tableXXX}
\begin{adjustbox}{width=0.47\textwidth}
\small
\begin{tabular}{cccccccccccc}


\hline
\multirow{3}{*}{\rotatebox[origin=r]{90}{Dataset}}  
  &\multirow{3}{*}{\rotatebox[origin=r]{90}{per ID}} & \multicolumn{3}{c}{Target set} & \multicolumn{2}{c}{Attack set} & Evaluation

\\ 
& &  train  & val  &  test  & member  &   non-member &   set \\

 &  &   $ID^1$ &$ID^1$  &$ID^1$ &$ID^1$ &$ID^2$ &$ID^{1,3}$ \\
 \\
 \hline
 
      \multirow{3}{*}{\rotatebox[origin=c]{90}{GREMM}}      \\
      \\ &   25    &  60  & 60  &   60  &   60  &   60 &   120 
               \\
       \\

 \hline

      \multirow{4}{*}{\rotatebox[origin=c]{90}{{\footnotesize Humpback}}} 
      \\
                        \\
                  & 25 &   90  & 90  &   90  &   90  &   90  &   180
                  \\
                  \\
                  \\

 \hline

      \multirow{3}{*}{\rotatebox[origin=c]{90}{NOAA}}  
                 &       \multirow{3}{*}{25} &   \multirow{3}{*}{127}    &   \multirow{3}{*}{127}    & \multirow{3}{*}{127}&    \multirow{3}{*}{127} &  \multirow{3}{*}{127}   &   \multirow{3}{*}{254}           \\
               \\

 \\

          \hline
         
\end{tabular}
\end{adjustbox}
\end{table}

As seen in Table~\ref{tableXXX}, the target set is split into a training, validation and test set, each set being composed of approximately one third of the pictures for each id.
The target training set is then used to train the target model while the validation and target test set are respectively used to validate the hyper-parameters and assess the accuracy of this model. 
Second, the attack set contains examples of non-members that are required to be able to train the attack model. 
In addition, pictures of the target validation are used as representatives of members.
Finally, the evaluation set contains non-members whose identities are different from the one used to built the attack model. 
The objective here is to assess the generalization power of the attack. 
Indeed, as the identities of the belugas in this set are different from the ones in the attack set, we are avoiding the situation in which the attack model overfits the attack set with respect to non-members.
Here, the target test set is used as examples of members for evaluating the success of the attack model. 
We balance each subset with the same number of individuals and samples to ensure that we can use the target validation set as members in our attack set and the target test set as members in the evaluation set.

\begin{figure}[h!]
  \centering
  \includegraphics[width=\linewidth]{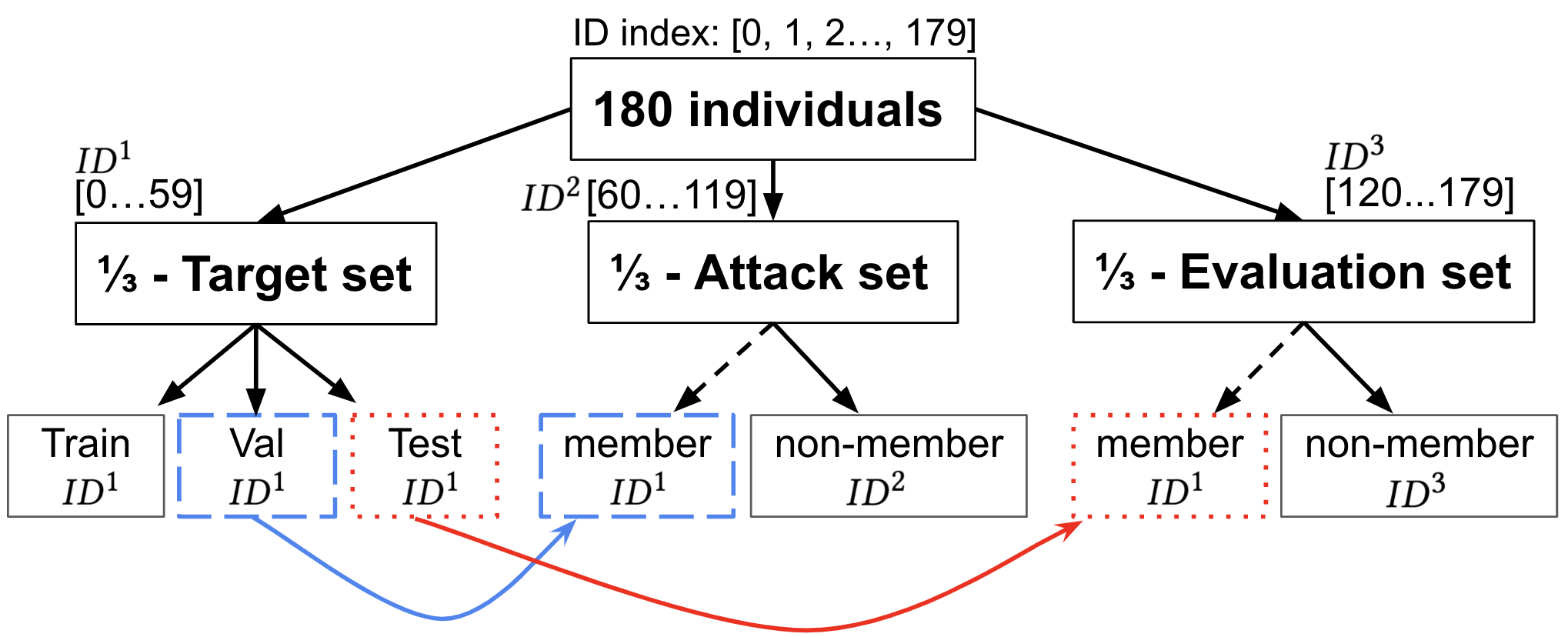}
  \caption{Dataset distribution. The main dataset is split in $\frac{1}{3}$ of individuals (\emph{e.g.}, 60 individuals in each subset: target set, attack set and evaluation set for GREMM dataset). The arrows indicate that the same individuals are re-used from one dataset to another.}
  \label{ExplaindatasetImage}
  
\end{figure}

Figure~\ref{ExplaindatasetImage} provides an example of the experimental configuration for the GREMM dataset, which contains 180 individuals.
Here $ID^1$, $ID^2$ and $ID^3$ represent subsets of pictures of whale individuals whose identities are totally different from one another. 
For instance, $ID^1$ contains individuals that belongs to the same IDs in target set but for which different samples (\emph{i.e.}, different pictures) are used to compose the training, validation and test set of the target model.
Thus, $ID^1$ individuals are considered as member in attack set (blue stroke rectangle) and evaluation set (red dotted rectangle). 
In contrast, $ID^2$ and $ID^3$ are individuals totally unknown by the target model (\emph{i.e.}, non-member). 
In a nutshell, the non-members of $ID^2$ are used to train the attack model while the ones of $ID^3$ are used to evaluate the attack performance of the attack model for IDs never seen before. 
This same schema is applied for Humpack and NOAA datasets, updating the numbers of individuals based on the information provided in Table~\ref{tableXXX}.

\subsection{Target and Attack Models}
\label{sec2}

We adopt two popular neural network architectures as our target model: ResNet50~\cite{resnet50} and DenseNet121~\cite{densenet121}. 

\begin{itemize}
\item \emph{ResNet50}. The ResNet50 architecture contains 50 layers and uses a stack of three layers with 1$\times$1, 3$\times$3, and 1$\times$1 convolutions as the building residual block. 
The three-layer residual block is designed as a bottleneck to enhance computational efficiency, in which the 1$\times$1 layers are responsible for reducing and then boosting (restoring) the dimensions, leaving the 3$\times$3 layer as a bottleneck with small input and output dimensions \cite{resnet50}. 
Batch normalization (BN)~\cite{bn} is applied after each convolution and before ReLU activation, and in addition the global average pooling (GAP)~\cite{lin}, is performed to form the final fully connected layer ($fc$) that contains the number of individuals of the respective dataset. 
After training, $fc$ outputs floating-point values, which corresponds to the predicted result.
\item \emph{DenseNet121}. 
The DenseNet architecture is designed around a simple connectivity pattern of dense blocks and transition layers. 
A dense block is a module containing many layers connected densely with feature maps of the same size. 
In a dense block, each layer obtains additional inputs from all preceding layers, and it passes on its own feature maps to all the subsequent layers.
The transition layer links two neighboring dense blocks and it reduces the size of the feature map through pooling. Compared with ResNet that connects layers through element-level addition, layers in DenseNet are connected by concatenating them at the channel level. 
Similar to ResNet, DenseNet uses a composite of three consecutive operations for each convolution: $BN$+$ReLU$+$convolution$.
\end{itemize}
These target models were trained in two different settings. 
\begin{itemize}
\item \emph{No-overfitting}. In this setting, the optimization algorithm of CNNs is Stochastic Gradient Descent (SGD), with a learning rate of 0.0001 and a weight decay of 0.5. 
The batch size is set to 32, the number of training epochs to 200 and finally the batch-norm and dropout (0.5) are used to reduce the overfitting level.
\item \emph{Overfitting}. We use the same hyperparameters setting as the no-overfitting but we remove the use of batch-norm, weight decay and dropout techniques to ensure that the model overfits. 
\end{itemize}

For neural network-based (\emph{i.e.}, Salem \emph{et al.}) and metric-based (\emph{i.e.}, Yeom \emph{et al.}) MIAs, the attack model use the same architectures and hyperparameters setting as the target model. 
For label-only attacks, we follow the implementation from Adversarial Robustness Toolbox (ART)~\cite{art}, which is an open source project that provides Python tools for developers to assess the robustness of machine learning models against security threats.

Similarly to previous work in the literature \cite{me1,salem2019,DBLP:conf/sp/ShokriSSS17,Choo2020}, we evaluate the attack performance using accuracy (\emph{i.e.}, attack success rate) as the main evaluation metric for both the original classification tasks and the MIAs.
We also evaluate the False Positive Rate (FPR), being aware that only the attack accuracy is not a sufficient measure to compute the success of the attack in open-set problems~\cite{carlini}.
As mentioned previously, the attack model is trained with the same architecture as the target model. 
However, in contrast to the standard setting of most of the MIAs in the literature~\cite{DBLP:conf/sp/ShokriSSS17,salem2019, Choo2020}, we assume that the non-members in the attack dataset come from a different distribution from the target dataset (\emph{i.e}, individuals never seen before in target dataset are part of attack dataset as seen in $D^2$).

\section{Results}
\label{sect_result}

In this section, we provide the results of our experiments and discuss the main findings that we can draw from them. 
More precisely in Section~\ref{comparisonLiteratureAttacks}, we compare the attack performance of proposed MIA algorithms for the different whale datasets. 
After in Section~\ref{generalization}, we discuss how the choice of the attack dataset and attack model’s architecture to evaluate the generalization power of the MIA.
Afterwards in Section~\ref{factor_overfittion_sampling_cnn}, we explore the influence of different factors, such as overfitting, on the attack's performance. 
Finally in Section \ref{reliabilityM}, we present the performance of our novel ensemble MIA for different real-world scenarios.

\subsection{Evaluation of MIAs}
\label{comparisonLiteratureAttacks}

\begin{figure}[h!]
  \centering
  \includegraphics[width=\linewidth]{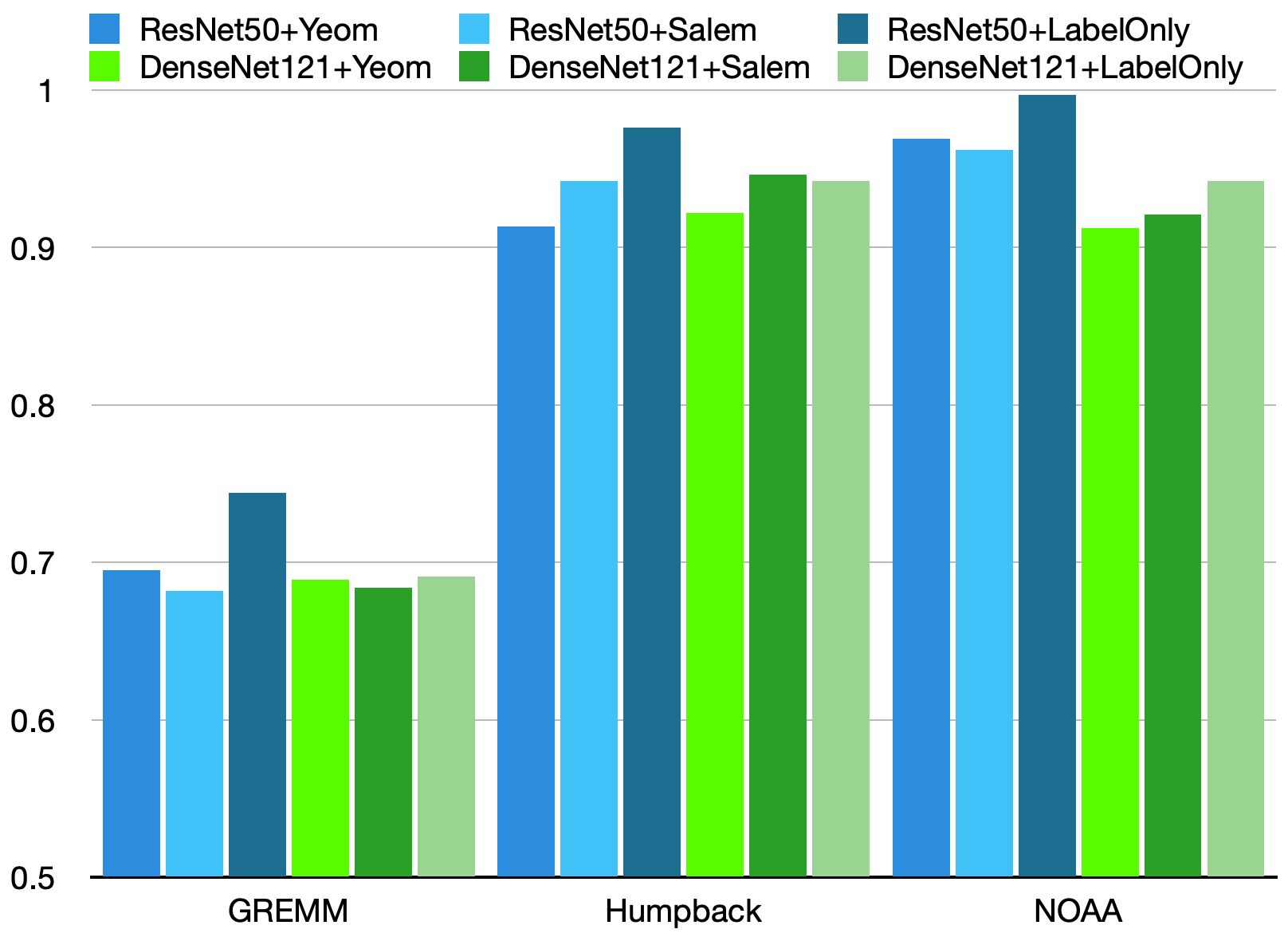}
  \caption{Accuracy of the different MIAs for different datasets and target model architectures. 
  As the evaluation set is balanced (\emph{i.e.}, composed of exactly half members and half non-members) a naïve attack model that produces a random prediction would have an accuracy of 0.5.}
  \label{Attackperformance}
  
\end{figure}

The attack performance of the MIAs was tested against different architectures for the target model, namely ResNet50 and DenseNet121.
Figure~\ref{Attackperformance} displays the performance of the different MIAs on different datasets. 
Overall, it can be observed that ResNet50+LabelOnly performs the best while ResNet50+Yeom and ResNet50+Salem have a lower performance. 
For example, on the NOAA dataset, the attack accuracy is 0.976 for ResNet50+LabelOnly against 0.913 for ResNet50+Yeom. 
This is expected as ResNet50+Yeom considers both confidence’s prediction and labels while ResNet50 +LabelOnly considers the predicted label’s correctness. 
More precisely, the predicted label’s correctness used by ResNet50 +Yeom is relatively coarse as many non-members are misclassified as members if the predicted label is correct. 
In contrast, LabelOnly MIA provides a finer-grained metric as it relies on the magnitude of perturbation to change the predicted label, which helps to further distinguish between members and non-members. 
However, LabelOnly requires a larger query budgets and computation costs than other attacks as it needs to query the target model multiple times and craft the adversarial perturbation to change the predicted label.

Table~\ref{timeconsum} presents a comparison of the training and discrimination time for the proposed MIAs. 
More precisely, the training time indicates the time required to train the attack model while discrimination time refers to the average computational time for a membership inference on a single data point. 
Nonetheless, metric-based MIAs can often achieve a performance that is not too far from the best attack. 
For instance, on the GREMM dataset, the attack performance of ResNet50+LabelOnly is 0.744 against 0.695 for ResNet50+Yeom.
Therefore, if the adversary has limited computation resources, metric-based MIAs may be a more appropriate choice than LabelOnly MIA.

\begin{table}[h!]
\centering

\caption{Computational time for training an attack model and running a single test image ("discrimination"). 
The numbers were obtained on the biggest dataset (NOAA), which contains 6350 samples for the attack set and 6350 samples for the evaluation set (\emph{i.e.}, members and non-members).
}\label{timeconsum}
\begin{tabular}{@{}|c|c|c|@{}}
\hline
  \multirow{2}{*}{Attack}&  Training Time &Discrimination Time  \\ 
\multirow{2}{*}{}  & (sec) & (sec) \\ 
\hline
Yeom \emph{et al.}~\cite{yeom2018} & 4.67 hr  & 0.66 s \\
Salem \emph{et al.}~\cite{salem2019} & 4.19 hr  & 0.62 s \\
Label-only~\cite{Choo2020}  & 6.41 hr  & 1.57 s \\

\hline
\end{tabular}

\end{table}

In addition, it seems that the attack performance can be improved by changing the model's architecture. 
For instance, on the Humpback dataset, the attack performance for DenseNet121+LabelOnly is 0.942 while an accuracy of 0.976 is achieved when the ResNet50 model's architecture is applied.
Thus, a more elaborated architecture is likely to boost the ability of the attack model to differentiate between members and non-members.
This is in line with the findings of recent studies in the literature~\cite{DBLP:conf/sp/ShokriSSS17, onThediff,8844607} that have shown that increasing the complexity of the model attacked is likely to increase the success of MIA due to the increase capacity of the model to memorize the training set.

Finally, we can observe that the success of the attacks is significantly lower for the GREMM dataset. 
Our intuition is that the hardness in discriminating beluga whale's in GREMM comes from the lack of discriminative characteristics. 
In particular, beluga individuals in the GREMM dataset are very similar to each other, forcing the attack model to misclassify non-members as members of the target model.
In contrast, individuals from the Humpback and NOAA datasets normally have very distinctive features (\emph{e.g.}, detailed features present in fins, marks and shapes). 
For instance, as seen in Figure~\ref{datasetImage} in comparison with Humpback and NOAA, GREMM individuals have no marks present in dorsal ridge or detailed tail, resulting in beluga whales being the hardiest species to attack.

\subsection{Generalization of the Attack}
\label{generalization}

Most previous works in the literature~\cite{me1,DBLP:conf/sp/ShokriSSS17} focused on the setting in which the adversary trains an attack model (of the same architecture as the target model) on an attack dataset that comes from the same distribution (\emph{i.e.}, members) as the target dataset. 
Generally, these works generate the target dataset and the attack dataset coming from the ``same distribution'' by splitting the original dataset into two parts. 
We depart from this assumption by creating an attack dataset composed for half of members and half of non-members totally different from the target dataset ($D^2$ in Figure~\ref{ExplaindatasetImage}) in the sense that even for the members the pictures used are different from the one used in the target dataset. 
More precisely, for the identity of a particular beluga contains in the target set, we have several pictures associated to it. 

When we built the different datasets as shown in Figure~\ref{ExplaindatasetImage}, we make sure that the pictures used for training the target model are different than the ones used for building the attack model or from the evaluation dataset.
In this situation, a successful attack means that the MIA will be able to generalize to new pictures of members as well as to new non-members. 
In particular, we want to be able to guarantee that the attack model has learned generic features rather than simply distinguishing in an overfitted manner the members and non-members of the attack set. 
In this situation, even when new individuals emerge over time, the attack model will be able to identify whether it is an individual known by the target model or not.  

In the following, we focus on the LabelOnly attack with ResNet50 architecture, which has shown the best performance in the experiments conduced and can handle the case in which the target and attack datasets come from different visual characteristics (\emph{e.g.}, beluga dorsal ridge in GREMM dataset, humpback tail in Humpback and beluga top-view in NOAA).
First, we analyze the situation in which we relax the assumptions of a same-distribution attack dataset and same architecture attack model. 
To realize, we evaluate whether a MIA is still effective against an attack dataset composed of non-members that are issued from a different dataset.

\begin{figure}[h!]
  \centering
  \includegraphics[width=\linewidth]{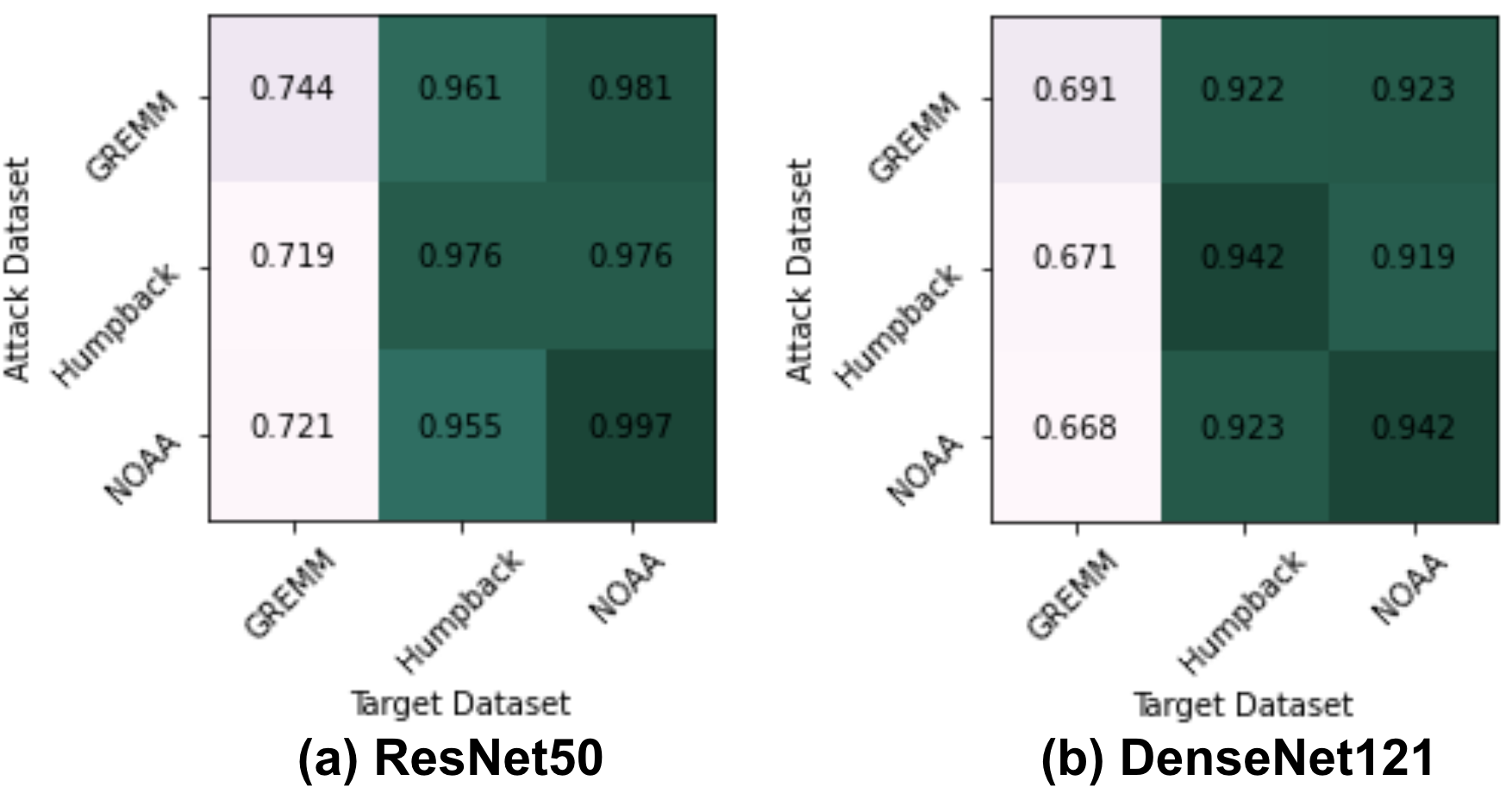}
  \caption{The performance of membership inference attack (LabelOnly) when the attack dataset comes from different distributions than the target dataset.}
  \label{Table_sets}
   
\end{figure}

\textbf{Attack performance on attack dataset coming from a different distributions.} So far, previous works~\cite{salem2019,DBLP:conf/sp/ShokriSSS17} have only considered the ``same distribution'' setting in which the attack dataset is based on images sampled from the same dataset. 
However, in reality, to construct an attack dataset for wild individuals, it might be the case that the system will unknown individuals that emerge over time. 
Figure~\ref{Table_sets} shows the MIA performance when the attack dataset contains non-members coming from a different distribution than the target dataset. 
In this situation, we can observe that the attack performance remains almost the same.
For instance, when the target and attack datasets both originated from GREMM, the attack performance is 0.744 while that attack is still effective (0.719 and 0.721) when the attack dataset originated respectively from Humpback and NOAA. 
Such observation indicates that we can relax the assumption of a same-distribution attack dataset.
In practice, this can have a big impact in the situation in which the target dataset is of limited size and we do not have the liberty to sacrifice some of its data to build the attack dataset.

The results obtained demonstrate that even if we add new individuals that have never been seen before by the attack model (\emph{e.g.}, from other dataset distributions), the attacks are still effective.
For instance, all attacks reach over 0.922 accuracy when the target dataset is Humpback and the attack dataset is GREMM or NOAA, even in the cases in which attack and target models have different model architectures. 
To the best of our knowledge, we are the first to quantify the generation power of MIAs with an attack dataset that is composed of half known members from the target set and another half of non-members totally unknown (\emph{i.e.}, from a different distribution). 

\begin{figure}[h!]
  \centering
  \includegraphics[width=\linewidth]{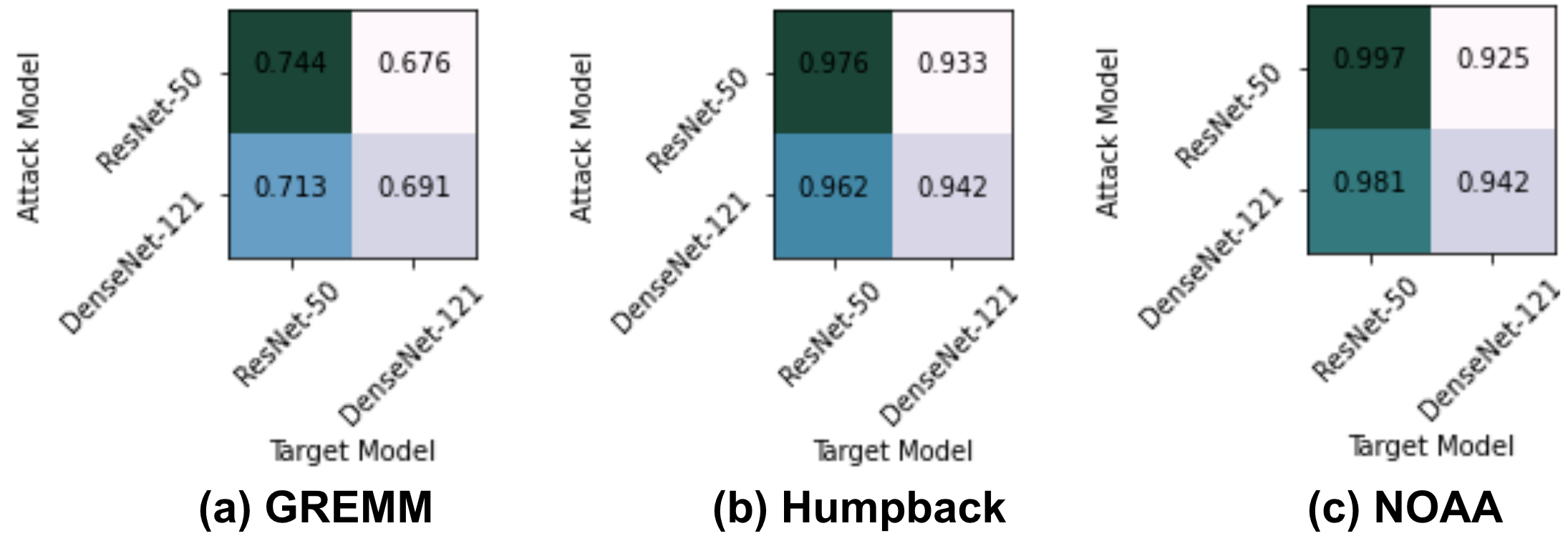}
  \caption{The performance of membership inference attack (LabelOnly) when the attack model has different architecture compared to the target model.}
  \label{MixArchicture}
   
\end{figure}

\textbf{Attack performance on different model's architecture.} 
Figure~\ref{MixArchicture} shows that the attacks are still effective even when the target and attack models’ architectures are different.
For instance, on the Humpback dataset (Figure~\ref{MixArchicture} b), the attack performance is 0.976 when ResNet50 is the model architecture for both target and attack models, and it decreases only to 0.962 when the attack model’s architecture changes to DenseNet121.
Such observation hints that we can relax the assumption that the attack model should necessarily follow the same-architecture as the target model.

\subsection{MIAs Influence Factors}
\label{factor_overfittion_sampling_cnn}

This section explores the factors that influence the success of MIAs in our setting. 
To realize this, we study how factors such as the overfitting level and cross-entropy distance distribution correlates with the attack performance. 
During our evaluation, we focus on the ResNet50+Salem, ResNet50+Yeom and ResNet50+LabelOnly attacks, as the former performs the best using only confidence information while the latter two perform the best when having access to both confidence and ground-truth label information.

\textbf{Difference between overfitting and no-overfitting.} 
The traditional way of training machine learning models normally aims at avoiding the overfitting phenomenon~\cite{avoidover, RAVOOR2020100289}. 
Indeed, the main concern about overfitting is that it occurs when the model performs well on the training data but generalizes poorly on unseen samples (\emph{i.e.}, test set). 
In the privacy domain, overfitting has also been shown to make the model more vulnerable to privacy attacks as it results in the model memorizing more information about the training set~\cite{8844607, DBLP:conf/sp/ShokriSSS17}. 

\begin{figure}[h!]
  \centering
  \includegraphics[width=\linewidth]{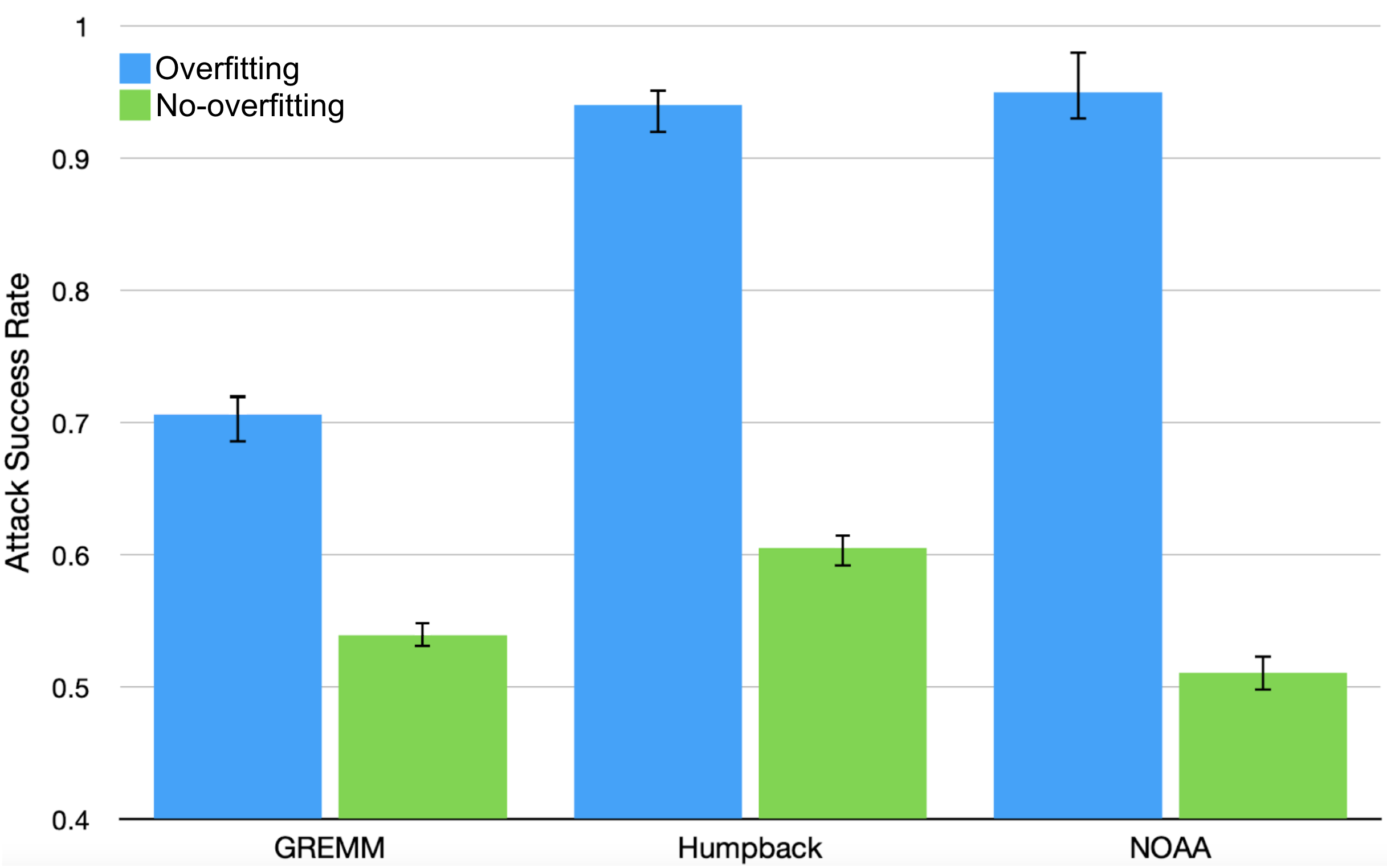}
  \caption{Success of MIAs in the overfitting vs no-overfitting settings. Note that we average the attack performance under different attacks for each dataset and show the standard deviations.}
  \label{OVERxNOover}
   
\end{figure}

In the following, we investigated how overfitting affects the performance of MIAs and more precisely whether overfitted models can more easily discriminate between known vs unknown individuals. 
When training the target models, we considered two different model settings: no-overfitting and overfitting as describe in Section~\ref{sec2}. 
The results of the experiments are summarized in Figure~\ref{OVERxNOover}.

As expected, models trained with overfitting displays a higher vulnerability against MIAs. 
For instance on GREMM, the best average attack accuracy for the original model (trai--ned with overfitting) was 0.706 while it was only 0.539 with no-overfitting, close to the 0.5 accuracy of a baseline random prediction. 
In terms of the utility of the target model with respect to MIAs, using overfitting always improves the MIA's generalization in all cases (\emph{i.e.}, for different datasets and model architectures).
Thus as expected, overfitting is effective at increasing the leak of information and can be leverage to discriminate more efficiently between members and non-members.

\textbf{Impact of the overfitting level.} 
As seen previously, the attack performance varies on the dataset and model considered. 
Previous works~\cite{yeom2018,overfff,DBLP:conf/sp/ShokriSSS17} have also explored how the level of overfitting impacts the success of privacy attacks.  
In a nutshell, the overfitting level of a given model can be defined by subtracting the testing accuracy from the training accuracy.
We report the training/testing accuracy on the classification tasks for overfitted and non-overfitted models in Table~\ref{NoWithoverfitted}.

\begin{table}[h!]
\centering

\caption{The performance for overfitting vs non-overfitting models on the original classification tasks for all three datasets. 
Both values of training accuracy and testing accuracy (in parenthesis) for different model architectures are reported. 
}\label{NoWithoverfitted}
\begin{tabular}{@{}cccccccccccc@{}}


\hline
\multirow{2}{*}{\rotatebox[origin=r]{90}{Data}} & \multicolumn{2}{c}{ResNet50} && \multicolumn{2}{c}{DenseNet121} 

\\

\cline{2-3} \cline{5-6}
 
&   \multirow{2}{*}{Overfitted}  &   \multirow{2}{*}{No}  & & \multirow{2}{*}{Overfitted}  &   \multirow{2}{*}{No}   
\\\\

\hline
 
      \multirow{4}{*}{\rotatebox[origin=c]{90}{GREMM}}      \\
      \\
       &     1.000 (0.282)  &   0.746 (0.406) & &   1.000 (0.222)  &   0.746 (0.356)      \\
       \\

\hline

      \multirow{4}{*}{\rotatebox[origin=c]{90}{Humpback}} 
      \\
      \\
                  &         1.000 (0.382)  &   0.893 (0.835) & &   1.000 (0.341)  &   0.893 (0.751)     \\
                  \\

\hline

      \multirow{3}{*}{\rotatebox[origin=c]{90}{NOAA}}  
                 &      \multirow{3}{*}{1.000 (0.371)}    & \multirow{3}{*}{0.872 (0.797)}& &   \multirow{3}{*}{1.000 (0.357)} &  \multirow{3}{*}{0.872 (0.737)}             \\
               \\
                
 \\

         \hline
         \end{tabular}

\end{table}

\begin{figure}[h!]
  \centering
  \includegraphics[width=\linewidth]{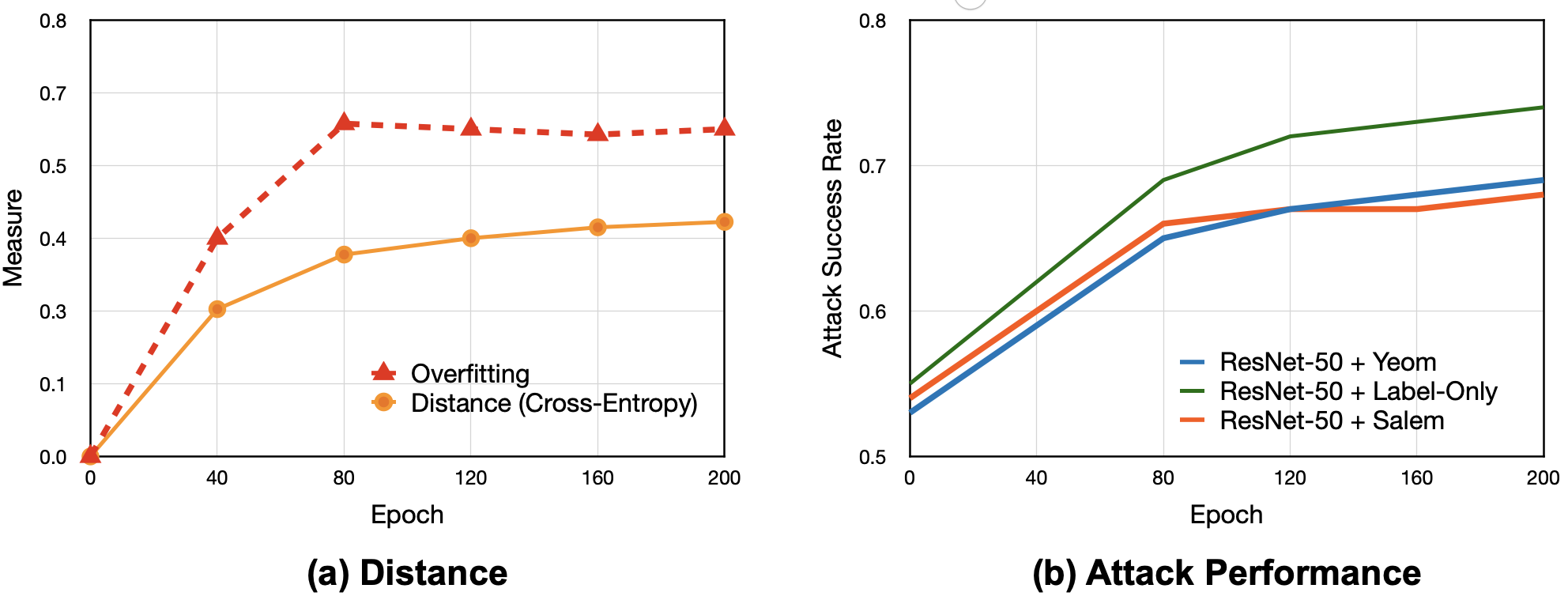}
  \caption{The distance in terms of cross-entropy and attack performance against the target model ResNet50 on the GREMM dataset under different numbers of epochs for model training.}
  \label{over_distance}
   
\end{figure}

Figure~\ref{over_distance} shows the correlation of the overfitting level correlation with the attack performance.
In particular, the MIAs vulnerability is associated with the increase of the overfitting level. 
For example, in Figure~\ref{over_distance}a, the overfitting level goes from 0 to 0.61 when the target model’s training epochs range from 0 to 80, which results in the attack success rate of the ResNet50+Label-only attack to vary from 0.55 to 0.69. 
This observation highlight the fact that the overfitting level contributes to the vulnerability of a model to MIAs. 
However, an unexpected outcome is that the attack performance is still increasing when the overfitting level stabilizes. 
As shown in Figure~\ref{over_distance}, when the overfitting level is around 0.6 (which corresponds to epochs ranging from 80 to 200), the attack performance still improves with the increase in the number of epochs. 
It shows that the overfitting level is not the only aspect related to MIA vulnerability. 
To address this issue, we additionally investigated the correlation between the distance in terms of cross-entropy between distributions for members and non-members and the vulnerability of the model to MIAs.

\textbf{Kullback–Leibler divergence.} 
We use the Kullback Leibler divergence (KL divergence)~\cite{kl} to measure the distance between the distributions of members and non-members and compute the cross-entropy of each sample. 
KL is a widely used metric to measure the distance of two probability distributions as seen in Equation~\ref{KLequation}. 

\begin{equation}
\mathcal{L}_{KL}(P,Q)=\sum_{x}P(x)\log\frac{P(x)}{Q(x)},
\label{KLequation}
\end{equation}
in which $P$ and $Q$ are two probability distributions on events. 
The loss function includes both the prediction loss and the KL divergence loss. 
From this, we can compute cross-entropy distributions for members and non-members and normalize them into probability distributions~\cite{kl}.
Cross-entropy loss is one of the most common loss functions used for classification tasks, and it is defined as:

\begin{equation}
\mathcal{L}_{CE}(y,p)=-\sum_{i=1}^ky_{i}\log p_{i},
\label{cross}
\end{equation}
in which $p$ is a vector that represents the confidence predictions of the sample over different pre-defined classes, with $k$ being the total number of classes. 
$y_i$ equals 1 only if the sample belongs to class $i$ and 0 otherwise while $p_{i}$ is the $i$-th element of the confidence posteriors.

More precisely, we computed the KL-divergence of the normalized cross-entropy distributions between members and non-members. 
Figure~\ref{over_distance}a shows the KL-divergence of cross-entropy distributions and the overfitting level under the target model trained with different epochs when the target model is ResNet50 trained on GREMM. 
We can see that the KL-divergence of cross-entropy is highly correlated with the attack performance. 
For example, in Figure \ref{over_distance}a, the KL-divergence of cross-entropy of the target model ranges from 0.0 to 0.40 when the epochs range from 0 to 120, with the attack success rate of ResNet50+LabelOnly varying from 0.55 to 0.72. 

More interestingly, from Figure \ref{over_distance}a and Figure \ref{over_distance}b, we can also see that there is a clear turning point after 120 epochs, in which both the KL-divergence and attack performance become stable. 
These results convincingly demonstrate that, compared to the overfitting level, KL-divergence of members’ and non-member’s cross-entropy has a higher correlation to the attack performance. 
Note that for LabelOnly attacks, we do not have confidence predictions but only the labels predicted by the target model. 
Thus, we can view the predicted label as the ground truth to calculate the cross-entropy loss instead of the KL-divergence loss in the distillation process.

    
\subsection{MIA Robustness and Performance of Ensemble MIA}
\label{reliabilityM}

Discrimination based on MIA might be impractical when the FPR is too high, which will lead to non-member samples being often erroneously predicted as members. 
We notice a smooth FPR for Humpback and NOAA datasets, with on average 0.03\% of FPR. 
This means that most of members and non-members are well discriminated for those datasets. 
In contrast, GREMM has a extremely high FPR, which can be decreased to 0.34\% using ResNet+LabelOnly attack under the overfitting influence.

\begin{figure}[h!]
  \centering
  \includegraphics[width=\linewidth]{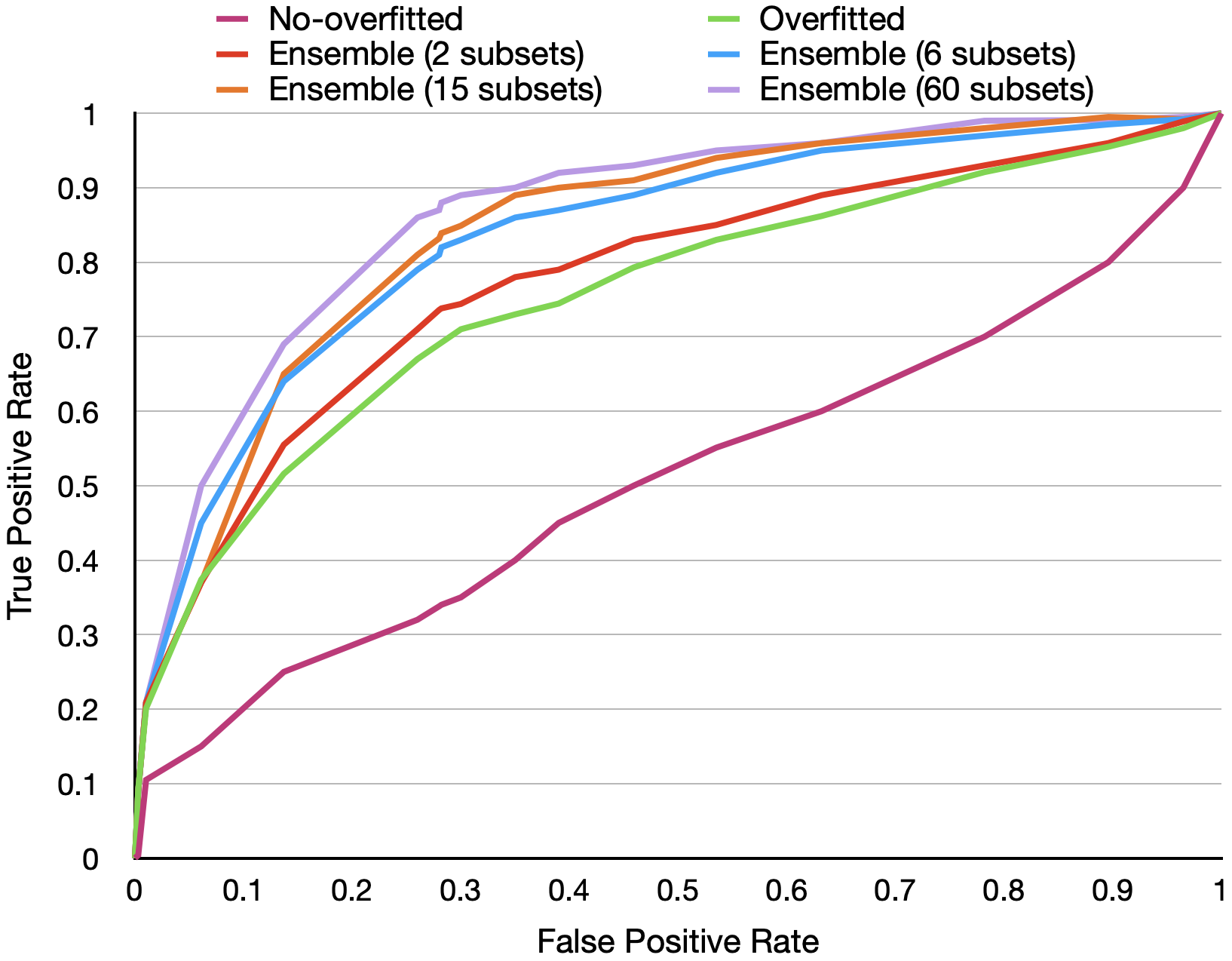}
  \caption{True positive versus false positive rates for different settings of MIAs.}
  \label{roc_MIA_GREEMimg}
   
\end{figure}

We have investigated the FPR obtained using the best proposed attack in Figure~\ref{roc_MIA_GREEMimg} for the GREMM dataset.
Some negative individuals are misclassified as positive due to hardness visual similarity between member and non-members in GREMM dataset. 
Interestingly, even the attack on an extremely overfitted model such as ResNet50 (green line) still suffers from high FPR.
This type of error makes the predicted membership signal unreliable, especially since most samples are non-members in real world applications. 
To reduce this, we have proposed the novel MIA ensemble approach (described in Section~\ref{sec_ensemble_mia}) to enhance the MIA performance while reducing the attack’s FAR.

While the gain in average attack success rate is modest, the success rate at low false-positive rates can be very high. 
For instance, looking at Table~\ref{Overffitingclasses}, we notice a variation in attack accuracy rate across different subsets of individuals. 
This suggests that there is a subset of examples that are easier to distinguish than others, which is also a phenomenon that has been observed in the literature on MIAs~\cite{liu2022}.
In our experiments, ensembles were design to explore the attack model using different sets of individuals. 
As seen in Figure~\ref{roc_MIA_GREEMimg} using ensemble composed by 2 subsets, we decreased the FPR to 0.35 while increasing the attack rate to 0.781. 
We further investigate whether as soon we increase the number of subsets the FPR might turns lower, which we observed as with 15 subsets we got 0.28 of FPR. 
The best results is when we create an attack model for each unique whale identity. 
The main insight behind the creation of attack models using unique individuals is that the member and non-member in the attack set are composed by two individuals being one known and the another unknown. 
The non-member is selected randomly from individuals never seen before to train the attack model. On this way, we guarantee the maximum overfitting level for unique individuals and merge the outputs of $M_{attack_{1}}$ to $M_{attack_{l}}$.
For instance, $M_{attack_{1}}$ to $M_{attack_{60}}$ ensemble with 60 outputs for GREMM dataset achieved 0.26 of FPR and 86\% in accuracy attack.
In fact, a high overfitting level using fewer individuals acts in synergy to increase the success rate of the attack while decreasing the FPR. 
In addition, ensemble MIA combines the output of each attack model to better discriminate similar individuals.





Finally, we have also performed additional experiments to investigate how the overfitting levels varies for different subset of individuals to observe whether the attack performance varies across subsets. 
For instance in Table~\ref{Overffitingclasses}, we have split the GREMM dataset respectively in two and six subsets. 
For example, with two subsets composed by individuals whose identities range between 0-29 and 30-59, the overfitting level is respectively 0.619 and 0.624.
This demonstrates that the membership leakage effect also varies among different individuals from the same dataset.

\begin{table}[h!]
\centering

\caption{The overfitting level in different attack subsets using (LabelOnly) when the target model is ResNet50 trained on GREMM. Class Index is the number of individuals used for each subset (\emph{e.g.}, 2 subsets containing 30 individuals each and 6 subsets with 10 individuals).}
\label{Overffitingclasses}
\begin{tabular}{@{}|l|c|c|@{}}
\hline
Class index   &         Overfitting Level     &  Subsets \\ \hline
0-29  & 0.619 & \multirow{2}{*}{2}             \\ 
30-59  & 0.624 &              \\ \hline
0-9  & 0.683 &   \multirow{6}{*}{6}           \\ 
10-19  & 0.604 &              \\ 
20-29  & 0.738 &              \\
30-39  & 0.752 &              \\
40-49  & 0.792 &              \\
50-59  & 0.724 &              \\
\hline
\end{tabular}
\end{table}

\label{experimentalSetup}

\section{Conclusion}
\label{sect_conc}

In this paper, we have performed MIAs against models trained on open-set datasets of whales with the objective of using it to be able to discriminate between known vs unknown individuals. 
More precisely, we have investigated three MIAs from the state-of-the-art using two popular model architectures as well as three whale benchmark datasets. 
Overall, the results obtained demonstrate that the combination of model architecture and MIA ResNet50+ LabelOnly performs the best and is able to discriminate members and non-members even when they have fine-grained visual similarity. 
We have shown that the assumption that the non-members should be from the same distribution can be relaxed. 
In particular, the non-members used to train the attack model could be taken from a different whale population without significantly impact the success of the discrimination.
Additionally, the results also highlight that the architecture of the attack model does not need to be similar to that of the target model. 
Finally, from the observation that the overfitting level in small subsets leads to a higher leak of information than larger subsets, we have proposed a novel approach called ensemble MIA. 
Ensemble MIA leads to an enhancement of 12\% in attack performance while  decreasing the FPR by 13\%. 

As future works, we would like to explore the use of white-box MIA to further improve the accuracy of the discrimination while reducing the FPR, in particular for the GREMM dataset.
We will also investigate how MIA-based approaches for discrimination compare to deep metric learning ones \cite{Bouma2019,Schneider2022}.
In addition while in this paper, our focus was on discriminating between the member and non-member whales, we plan to integrate the proposed MIA into a full pipeline for beluga whale re-id that we plan to open source. 
Finally, we hope that our work, in which we leverage on privacy attacks to address practical challenges encountered in animal ecology, will foster further research at the crossing of these two domains.

\bibliographystyle{abbrv}
\bibliography{vldb_sample}


\end{document}